\documentclass[10pt,twocolumn,letterpaper]{article}
\usepackage{authblk}
\usepackage[symbol]{footmisc}

\usepackage{cvpr}              

\usepackage{graphicx}
\usepackage{amsmath}
\usepackage{amssymb}
\usepackage{booktabs}

\usepackage[pagebackref,breaklinks,colorlinks]{hyperref}

\usepackage[capitalize]{cleveref}
\crefname{section}{Sec.}{Secs.}
\Crefname{section}{Section}{Sections}
\Crefname{table}{Table}{Tables}
\crefname{table}{Tab.}{Tabs.}

\def\cvprPaperID{*****} 
\def\confName{CVPR}
\def\confYear{2023}

\begin{document}

\title{Integrating Holistic and Local Information to Estimate Emotional Reaction Intensity}

\date{}

\author[1]{\textbf{Yini Fang$^\ast$}}
\author[1]{\textbf{Liang Wu$^\ast$}}
\author[2]{\textbf{Frederic Jumelle}}
\author[1]{\textbf{Bertram Shi}}

\affil[1]{Hong Kong University of Science and Technology}
\affil[2]{Bright Nation Limited}

\affil[ ]{\textit {\{yfangba, liang.wu\}@connect.ust.hk, f.jumelle@brightnationlimited.com, eebert@ust.hk}}

\maketitle

\begin{abstract}
Video-based Emotional Reaction Intensity (ERI) estimation measures the intensity of subjects' reactions to stimuli along several emotional dimensions from videos of the subject as they view the stimuli. We propose a multi-modal architecture for video-based ERI combining video and audio information. Video input is encoded spatially first, frame-by-frame, combining features encoding holistic aspects of the subjects' facial expressions and features encoding spatially localized aspects of their expressions. Input is then combined across time: from frame-to-frame using gated recurrent units (GRUs), then globally by a transformer. We handle variable video length with a regression token that accumulates information from all frames into a fixed-dimensional vector independent of video length. Audio information is handled similarly: spectral information extracted within each frame is integrated across time by a cascade of GRUs and a transformer with regression token. The video and audio regression tokens' outputs are merged by concatenation, then input to a final fully connected layer producing intensity estimates. Our architecture achieved excellent performance on the Hume-Reaction dataset in the ERI Esimation Challenge of the Fifth Competition on Affective Behavior Analysis in-the-Wild (ABAW5). The Pearson Correlation Coefficients between estimated and subject self-reported scores, averaged across all emotions, were 0.455 on the validation dataset and 0.4547 on the test dataset, well above the baselines. The transformer's self-attention mechanism enables our architecture to focus on the most critical video frames regardless of length. Ablation experiments establish the advantages of combining holistic/local features and of multi-modal integration. Code available at \url{https://github.com/HKUST-NISL/ABAW5}.

\end{abstract}
\footnotetext[1]{The first two authors contributed equally in the paper.}

\section{Introduction}
\label{sec:intro}

Attention to human affective behaviour analysis has increased in recent years, due to the multitude of its applications to fields such as robotics, human computer interaction (HCI) and psychology. The fifth Competition on Affective Behaviour Analysis in-the-wild (ABAW5)  focuses on affective behaviour analysis in-the-wild. Solutions to its proposed challenges can be used to create systems that can understand people's feelings, emotions and behaviours, as well as machines and robots that can serve as `human-centered' digital assistants. 

The Emotional Reaction Intensity (ERI) Estimation Challenge in ABAW5 addresses one of the contemporary affective computing problems using the Hume-Reaction dataset \cite{christ2022muse}, where subjects react to a wide range of emotional video stimuli while being observed by a webcam in their homes. After viewing each video, subjects self-annotate the intensity (over a range from 1 to 100) of their emotional reactions to it along seven dimensions (Adoration, Amusement, Anxiety, Disgust, Empathetic Pain, Fear, and Surprise).

Many previous approaches to video-based ERI estimation follow a similar pattern for representing video information, using features taken from among the final layers of a deep network that has been pre-trained on a large dataset, such as AffectNet \cite{mollahosseini2017affectnet}. AffectNet is annotated for holistic judgements of emotional facial expressions: categorization into one of eight emotion classes (neutral, happy, angry, sad, fear, surprise, disgust, contempt) and estimates of the intensity of valence and arousal.Thus, features extracted from later layers encode holistic judgements made by integrating information across the entire face. We hypothesize that such features may not encode more subtle, spatially localized changes in facial geometry, that may be important for estimating reaction intensity when subjects are simply viewing stimuli by themselves, 

One way of labelling spatially localized changes is the Facial Action Coding System \cite{ekman1978facial}, which defines a set of action units (AUs) that taxonomizes the configuration of small groups of human facial muscles by their appearance on the face. Action units are spatially localized. For example, facial units include ``inner brow raiser",  ``chin raiser", ``lip corner puller", etc. There are 28 main AUs, but most databases are annotated for only a subset (e.g., 12 or 17) of these. Annotations can either be binary (indicating occurrence) or continuous (indicating intensity). As AUs are spatially localized, we hypothesize that they may encode information that is complementary to that encoded by later layers of a deep net trained on AffectNet, yet that is critical for ERI.  

There are two main issues we address here in estimating a single emotional intensity along each emotional dimension in response to an entire video. First, videos can be of varying length. For example, in the Hume-Reaction dataset, videos range in length from 9.9 seconds to 14.98 seconds. Second, many of the frames within the video are images of the subject with a neutral expression, as it is often that only isolated moments in the video evoke changes in expression. Unfortunately, these moments are not known in advance. The unpredictability and sparsity of these events suggest that some common methods for extracting fixed length representations of a variable length video for input to a final classifier, such as randomly or uniformly sampling a fixed number of frames, or by average pooling across all frames in the video, are not appropriate, as they may miss or dilute the impact of critical frames. We address this problem through the use of a transformer architecture, where each frame corresponds to an input token, but we also add a regression token similar to the class token used in the Visual Transformer \cite{dosovitskiy2020image}, which gathers information from all frames with weighting by an attention value. Using the output of the regression token as input to the final classification stage ensures a fixed dimensional representation that can be computed by combining information from all frames of a variable length video. The attention weighting avoids the problem of dilution by many frames containing neutral expressions, which might be problematic, especially in longer videos. 

This paper proposes a  multimodal architecture for ERI estimation that includes the two approaches outlined above. Our architecture uses both video and audio features, extracted from pre-trained networks (for video) or handcrafted algorithms (for audio). To the best of our knowledge, this is the first time that separate representations of holistic (AffectNet based) and local (AU-based) visual features have been combined for ERI estimation. Temporal information is extracted in two stages, first by a GRU, which operates sequentially from video or audio frame-to-frame, and second by a transformer with regression token, which combines information globally and in parallel. Our network adopts a late fusion architecture, where video and audio information is combined after separate temporal aggregation stages, just before the final classification layer. However, our architecture can be easily modified for early fusion, a promising direction for future development. Our architecture achieved excellent performance compared to baseline, outperforming the multimodal baseline by 91.02\% on validation set and by 124.1\% on test set. 

\section{Related Work}
\label{relatedwork}
Emotional Reaction Intensity Estimation with the same dataset has also been presented in a Hume-Reaction MuSe 2022 \cite{christ2022muse} sub-challenge. The FaceRNET \cite{kollias2023facernet} was the best-performing model. It uses a CNN-RNN model to capture spatial-temporal correlations of all the frames in the video, with a routing mechanism on top. Wang et al. \cite{wang2022emotional} proposed a spatiotemporal transformer architecture for dynamic facial representation learning and a multi-label graph convolutional network for emotion dependency modelling. ViPER \cite{vaiani2022viper} leverage the video's multimodal nature, and propose a transformer-based model to combine video frames, audio recordings, and generated textual annotations. These work neglect the importance of Action Unit feature in ERI estimation, and can be further improved by incorporating AU occurrence and intensity. The Hume-Reaction MuSe 2022 challenge also reported results from a baseline system that used Py-Feat \cite{jolly2021py} to detect  the occurrence of 20 different AUs in each frame, and used these 20 binary values as the input to a Long Short-Term Memory LSTM-RNN. However, they did not investigate the addition of other visual features besides AUs, as we describe here. 


\begin{figure*}[t]
\includegraphics[width=16cm]{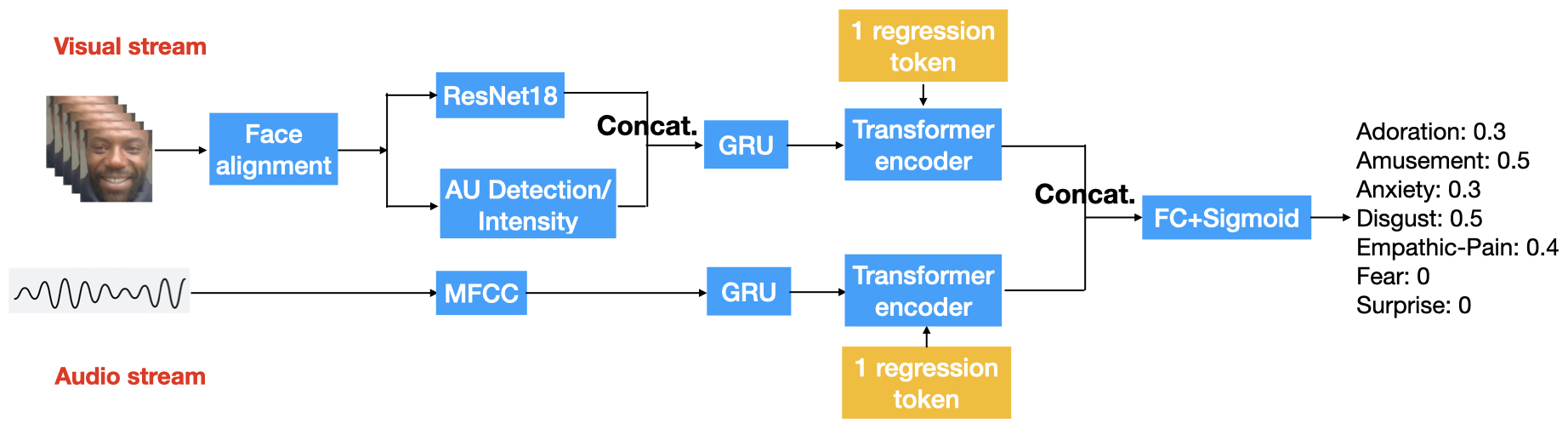}
\centering
\caption{Our architecture consists of two steams: a video stream and an audio stream. Visual feature includes holistic feature from a pretrained ResNet18 and local feature from AU detection/intensity estimation. Audio feature is extracted by MFCC. Each stream comprises GRU and transformer encoder to integrate information temporally. A learnable regression token is added to the input to the transformer encoder. The regression tokens from two streams are concatenated and fed into fully connected layer and Sigmoid for the final prediction. ``FC" stands for Fully Connected Layer. ``Concat." stands for concatenation.}
\label{model}
\end{figure*}

\section{Problem Formulation}
We denote by $\mathbb{V}$ and $\mathbb{A}$ the visual and audio stream of a reaction video, and by $\mathbb{Y} = \{y_1, ..., y_7\} $ seven emotional reaction intensities, representing Adoration, Amusement, Anxiety, Disgust, Empathic Pain, Fear, and Surprise, respectively.

Our objective is to predict 7 emotional reaction intensities $\mathbb{X} = \{x_1, ..., x_7\} $ given $\mathbb{V}$ and $\mathbb{A}$ using our proposed architecture $\mathcal{M}$, i.e., 
\begin{equation}
\mathbb{X} = \mathcal{M} ( \mathbb{V},  \mathbb{A}),
\end{equation}
so that the loss $\mathcal{L}(\mathbb{X}, \mathbb{Y})$ is minimal.

In the next section, we will explain $\mathcal{M}$ and $\mathcal{L}$ in detail.

\section{Methodology}

Figure \ref{model} shows our system's architecture. Given a video, our system outputs 7 emotional reaction intensities. Our architecture is dual stream: including a video stream and an audio stream. The video-stream is further separated into dual streams: one for holistic feature extraction (a ResNet18 network trained on AffectNet) and one for local feature extraction (AU detection/intensity estimation by OpenFace). Dual stream features are combined by concatenation before being input to the next stage. In the video stream, we extract information spatially for each frame first, then integrate across time by a cascade of a Gated Recurrent Unit (GRU) and a transformer encoder. In the audio stream, we extract spectral information from frames of length $\sim$30ms, then follow a similar temporal integration architecture as used in the video stream. Finally, we fuse visual and audio features by concatenation. A fully connected layer merges multimodal information to output seven emotional reaction intensities. We will elaborate these steps in details in the following.

\subsection{Visual Feature Extraction}

\paragraph{Data Preprocessing}

To eliminate variability due to head motion, which interferes with facial expression recognition, we apply face alignment using the OpenFace Toolkit \cite{baltrusaitis2018openface}, which detects 68 facial landmarks. We use them to align the face by linear warping followed by cropping of the face region. Then we resize the cropped face images to  $224 \times 224$ pixel resolution.

\paragraph{Holistic Spatial Features} We use a ResNet18 network \cite{he2016deep} pretrained on AffectNet \cite{mollahosseini2017affectnet} to extract holistic information about the general overall impression of the face. AffectNet is a large facial expression dataset consisting of 0.4 million images, designed for supervised facial analysis tasks. We use the 512-dimensional feature vector immediately before the final classification layer, which contains holistic information about the facial expression.

\paragraph{Local Spatial Features}
We use the OpenFace AU detection model \cite{baltruvsaitis2015cross} to extract both the occurrence and intensity of 17 AUs (e.g., AU1, AU2, AU4, AU6, AU7, AU10, AU12, AU14, AU15, AU17, AU23, AU25). This results in a 34-dimensional AU feature vector for each frame. This model is based on appearance (Histograms of Oriented Gradients) and geometry features (shape parameters and landmark locations). They propose a person-specific normalization approach, which allows the model to generalize well on other datasets.

\paragraph{Visual Feature Fusion}
We fuse holistic and local spatial features by concatenation, resulting in a 546-dimensional visual feature vector.

\subsection{Audio Feature Extraction} 
Mel-frequency Cepstral Coefficients (MFCC), the overall shape of the spectral envelope, have been widely used in speech recognition tasks. We extract them from $\sim$30ms frames sampled at 16kHz and with a stride of $\sim$16ms with the Python Librosa toolkit and use them as the basis of our audio feature representation. From a set of 32 MFCCs per frame, we obtain a sequence of 1024-dimensional audio feature vectors by combining features from 32 adjacent frames.


\subsection{Temporal Integration}
Each branch cascades a GRU and a transformer encoder to integrate information from a variable number of frames per video into a final fixed-dimensional feature vector, which is used for the final prediction.

\paragraph{GRU} The GRU operates sequentially, frame to frame, to capture short- and long-term temporal correlations in the video. 

\paragraph{Transformer Encoder} A Transformer encoder stacks several blocks. Each block has the same architecture, containing a multi-head attention mechanism followed by a fully-connected feed-forward network. Each frame corresponds to one token. We also add a regression token, with a set of learned embedding parameters. Through the self-attention mechanism, the regression token gathers information from the frames in the video. Only the regression token output after the last stage of the transformer encoder is passed to the next stage. 



\paragraph{Readout function} We input the concatenation of the video and audio regression token outputs into the final readout layer, consisting of a fully connected linear layer followed by a logistic sigmoid output layer, which restricts the output into [0, 1]. For each video sample $n$, our model outputs 7 scores $\mathbb{X}^n = \{x_1^n, ..., x_7^n\} $.

\subsection{Training} We freeze the weights in ResNet18 and AU detection model for the sake of training speed. We train the rest of the model using L2 loss:
\begin{equation}
\mathcal{L}(\mathbb{X}, \mathbb{Y}) = \frac{1}{7 \times N} \sum_{i=1}^{7} \sum_{n=1}^{N} (x_i^n - y_i^n)^2,
\end{equation}
where $N$ is the total number of video samples.

\section{Experimental Results}

\begin{figure*}[t]
\includegraphics[width=15cm]{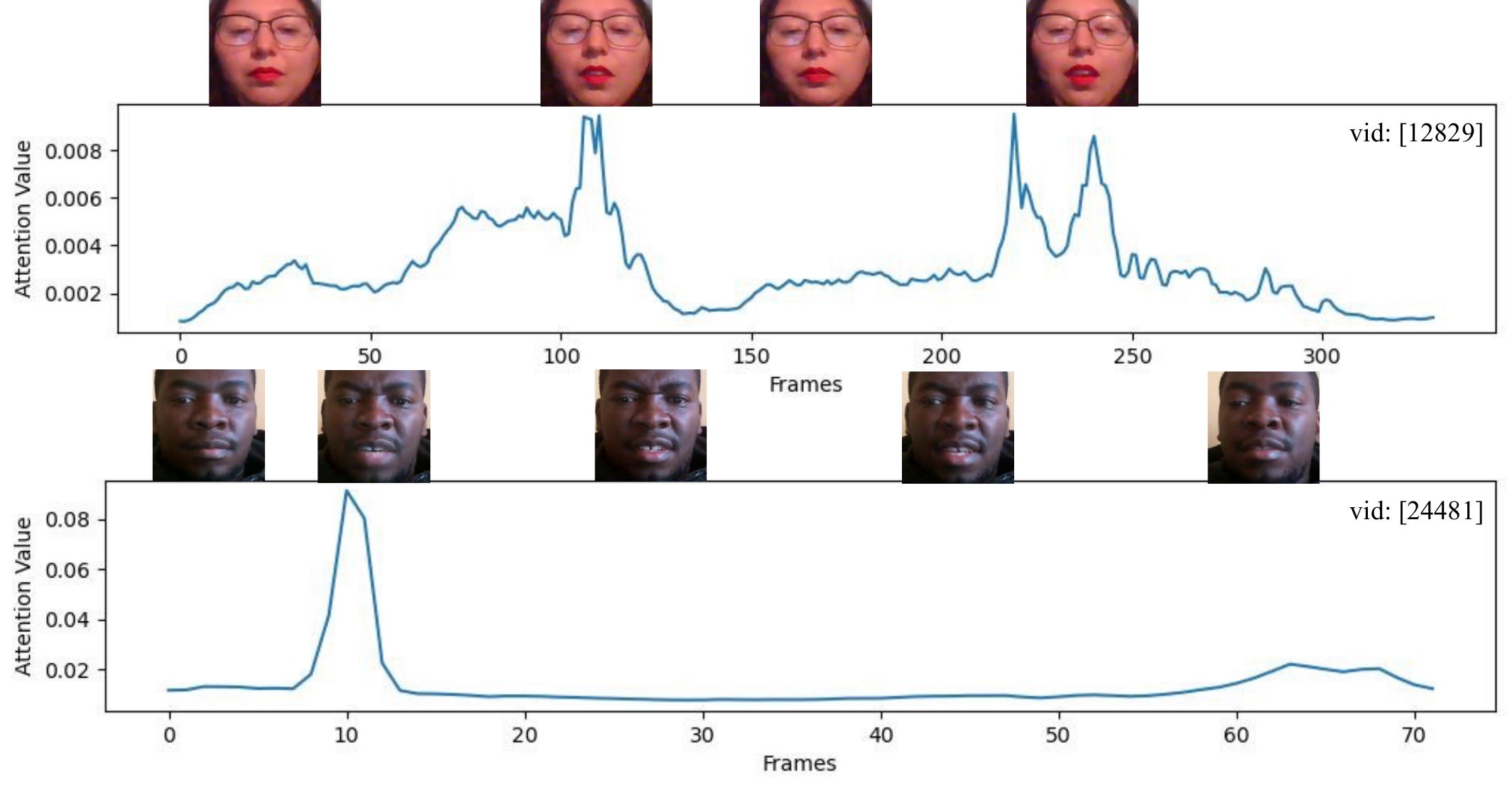}
\centering
\caption{Attention from the regression token to each frame in the video (averaged across all heads).}
\label{fig:attn}
\end{figure*} 

\subsection{Dataset}
The Hume-Reaction dataset consists of 25,067 videos from 2,222 subjects. Subjects are from two cultures, 1,084 from South Africa and 1,138 from the United States, aged from 18.5 to 49.0 years old. The reaction videos vary in terms of resolution, FPS, background, and lightning noise conditions. The average video length is 11.62 seconds, and the average number of frames per video is 248.8 frames.

The dataset is partitioned into three sets: 15,806 for training set, 4,657 for validation set, and 4,604 for testing set. 

\subsection{Evaluation Metric}
Pearson’s correlations coefficient (PCC) is used for the evaluation metric. We calculate the average PCC across the 7 emotional reactions intensities:
\begin{equation}
\mathcal{P}_{E R I}=\frac{\sum_{i=1}^{7} \rho_{i}}{7}.
\end{equation}
And for each emotion, $\rho_i$ is defined as:

\begin{equation}
\rho_i=\frac{\sum_{n=1}^N(x_i^n-\bar{x}_i)(y_i^n-\bar{y}_i)}{\sqrt{\sum_{n=1}^N(x_i^n-\bar{x}_i)^2}\sqrt{\sum_{n=1}^N(y_i^n-\bar{y}_i)^2}},
\end{equation}
where $\bar{x}$ and $\bar{y}$ are the mean of predictions and labels.


\subsection{Baselines}
We compare our model with the work that we have mentioned in Sec. \ref{relatedwork} and the provided audio/visual baselines. The audio baseline uses DEEPSPECTRUM \cite{amiriparian2017snore} to extract features from the audio signal. The visual baseline uses the feature vector from the last layer of a ResNet50 \cite{he2016deep} trained on VGGface2 \cite{cao2018vggface2}. Then LSTM-RNN is used to process these features and output the prediction.

\subsection{Implementation Settings}
We use a 2-layer GRU with the size of a hidden layer being 256. The number of transformer encoder blocks and the heads in the multi-head attention layer are both set to 4. 
A dropout probability of 0.2 was adopted for the transformer encoder. All our models are implemented using PyTorch and trained on a single GeForce 3070 GPU.

For the training parameters, we set an initial learning rate as 1e-4 which decays every 10 epochs by a factor of 0.5. AdamW optimizer with a weight decay of 0.5 is applied as the optimizer. 

Due to the varying quality and noise of the video, there are 1\% frames where no face is detected. We discard these frames. We also discard videos that have less than 50 valid images in the training. In the testing, the prediction of such an invalid video is set to be the average seven scores from other valid videos, in order to remove its impact on the PCC metrics. 

\subsection{Results}
Table \ref{result} shows performance comparison on the validation and test sets. Our model achieved 0.455 on the validation dataset and 0.4547 on the test data set, surpassing all the existing methods in the table on both validation and test sets. Compared to multimodal baseline, our model outperforms it by 91.02\% on validation set and by 124.1\% on test set, proving the effectiveness of adding AU features.


Figure \ref{fig:attn} shows the attention from the regression token to each frame of one long video sample and one short video sample in the validation set. The peaks in the attention curves show that the attention mechanism can identify frames that have the most informative expression changes. The peaks are quite sparse in the video, indicating the importance of using all frames for the estimation, and the ability of the network to avoid dilution of the information in key frames by neutral expressions present in most of the frames. 



\begin{table}[t]
\centering
\caption{PCC Comparison}
\label{result}
\begin{tabular}{|l|l|l|}
\hline
Methods         & Val & Test \\ \hline
\begin{tabular}[c]{@{}l@{}}Audio baseline\\ (DEEPSPECTRUM) \cite{christ2022muse} \end{tabular}  & 0.1087   & 0.0741      \\ \hline
\begin{tabular}[c]{@{}l@{}}Visual baseline\\ (VGGFACE2) \cite{christ2022muse} \end{tabular} & 0.2488 & 0.1830        \\ \hline
Visual baseline (FAU) \cite{christ2022muse} & 0.2840 & 0.2801 \\ \hline
Multimodal baseline \cite{christ2022muse} & 0.2382 & 0.2029 \\ \hline
Former-DFER+MLGCN \cite{wang2022emotional} & 0.3454 & N/A
\\ \hline
ViPER \cite{vaiani2022viper} & 0.3025 & 0.2970
\\ \hline
FaceRNet \cite{kollias2023facernet}  & 0.3590 & 0.3607 \\ \hline
ResNet18 \cite{li2022hybrid} & 0.3893 & N/A \\ \hline
ResNet18+DEEPSPECTRUM \cite{li2022hybrid} & 0.3968 & N/A \\ \hline
Ours            & \textbf{0.4550}   & \textbf{0.4547}       \\ \hline
\end{tabular}
\end{table}

\section{Ablation Study}
To study the effect of different types of features (e.g., holistic ResNet18 feature, local AU feature, and audio feature), we conducted experiments with different combinations on the validation set, shown in Table \ref{feat}. Using all the features achieved the best performance. We can also see that incorporating AU features lead to a boost in performance.

\begin{table}[t]
\centering
\caption{Ablation study on different feature types}
\label{feat}
\begin{tabular}{|l|l|}
\hline
Combination    & Validation \\ \hline
Only audio & 0.299         \\ \hline
Only AU & 0.342          \\ \hline
Only ResNet18 & 0.362          \\ \hline
ResNet18 + AU & 0.377          \\ \hline
ResNet18 + audio   & 0.438         \\ \hline
ResNet18 + AU + audio  & 0.455          \\ \hline
\end{tabular}
\end{table}

We also compared the effect of different types of AUs in Table \ref{au}. We can see that using only AU intensity is slightly better than using only occurrence, while combining both of them yields the best outcome.


\begin{table}[t]
\centering
\caption{Ablation study on AU features}
\label{au}
\begin{tabular}{|l|l|}
\hline
Combination    & Validation \\ \hline
AU occurrence &   0.452   \\ \hline
AU intensity &    0.454   \\ \hline
AU occurrence + intensity & 0.455          \\ \hline
\end{tabular}
\end{table}

\section{Conclusion}
This paper describes a multimodal architecture for Emotional Reaction Intensity estimation. Our results demonstrate the efficacy of including both holistic and spatially localized information from face images for this task. Our use of a transformer encoder with a regression token enables the network to handle videos with varying length in a consistent manner and to handle the sparsity of frames indicating emotional reactions within both long and short videos. Our experimental results indicate that the performance of our architecture significantly outperforms the baselines. We also conducted ablation experiments to verify the importance of multiple visual cues and multiple sensory modalities. Our code implementation is available at \url{https://github.com/HKUST-NISL/ABAW5}.

{\small
\bibliographystyle{ieee_fullname}
\bibliography{PaperForReview}
}

\end{document}